\documentclass[%
 aip,
 amsmath,amssymb,
 reprint,%
]{revtex4-1}

\usepackage[version=3]{mhchem} 
\usepackage{graphicx}
\usepackage{dcolumn}
\usepackage{bm}
\usepackage{float}
\usepackage{hyperref}

\usepackage[utf8]{inputenc}
\usepackage[T1]{fontenc}
\usepackage{mathptmx}
\usepackage{etoolbox}
\usepackage{siunitx}
\usepackage{amsmath}
\DeclareSIUnit\angstrom{\text {Å}}

\DeclareMathOperator*{\argmin}{arg\,min}

\makeatletter
\def\@email#1#2{%
 \endgroup
 \patchcmd{\titleblock@produce}
  {\frontmatter@RRAPformat}
  {\frontmatter@RRAPformat{\produce@RRAP{*#1\href{mailto:#2}{#2}}}\frontmatter@RRAPformat}
  {}{}
}%
\makeatother
\begin{document}

\preprint{AIP/123-QED}

\title[]{Neural Network Predicts Ion Concentration Profiles under Nanoconfinement}
\author{Zhonglin Cao}
\affiliation{Department of Mechanical Engineering, Carnegie Mellon University, Pittsburgh, PA 15213}
\author{Yuyang Wang}%
\affiliation{Department of Mechanical Engineering, Carnegie Mellon University, Pittsburgh, PA 15213}
\author{Cooper Lorsung}%
\affiliation{Department of Mechanical Engineering, Carnegie Mellon University, Pittsburgh, PA 15213}

\author{Amir Barati Farimani}
\affiliation{Department of Mechanical Engineering, Carnegie Mellon University, Pittsburgh, PA 15213}
\affiliation{Department of Chemical Engineering, Carnegie Mellon University, Pittsburgh, PA 15213}
\affiliation{Machine Learning Department, Carnegie Mellon University, Pittsburgh, PA 15213}
 \email{barati@cmu.edu}

\begin{abstract}
Modeling the ion concentration profile in nanochannel plays an important role in understanding the electrical double layer and electroosmotic flow. Due to the non-negligible surface interaction and the effect of discrete solvent molecules, molecular dynamics (MD) simulation is often used as an essential tool to study the behavior of ions under nanoconfinement. Despite the accuracy of MD simulation in modeling nanoconfinement systems, it is computationally expensive. In this work, we propose neural network to predict ion concentration profiles in nanochannels with different configurations, including channel widths, ion molarity, and ion types. By modeling the ion concentration profile as a probability distribution, our neural network can serve as a much faster surrogate model for MD simulation with high accuracy. We further demonstrate the superior prediction accuracy of neural network over XGBoost. Lastly, we demonstrated that neural network is flexible in predicting ion concentration profiles with different bin sizes. Overall, our deep learning model is a fast, flexible, and accurate surrogate model to predict ion concentration profiles in nanoconfinement.
\end{abstract}

\maketitle

\section{Introduction}
Nanoconfinement defines a state when fluid is confined in nanoscale spaces such as nanotube and nanochannel\cite{sun2020nanoconfined}. Compared with bulk condition, nanoconfined fluid such as water exhibit anomalous viscosity\cite{thomas2008reassessing, wu2017wettability}, diffusion coefficient\cite{barati2011spatial, wang2018layered, cao2022diffusion}, phase\cite{barati2016existence, giovambattista2009phase}, dynamics\cite{mondal2020different}, and density\cite{hu2010water, barati2011spatial, pan2020nanoconfined}. These behaviors become more exotic once we add ions to water. When ionic solution is confined by charged nanochannel walls, electrical double layer (EDL) is formed by the ions at the solid-liquid interface\cite{rojano2022effect, qiu2016ionic}. EDL is the basis for the development of advanced energy storage devices such as capacitors and batteries\cite{tan2022nanoconfined}. EDL can be engineered by surface modification to achieve the maximum charge storage capacity. Ions in EDL move with the solvent molecules when external electrical field is applied, resulting in electro-osmotic flow in the nanochannel\cite{qiao2004charge}. Electro-osmotic flow is a widely-used fluidic transport mechanism as it is easily controllable and scalable\cite{qiao2004charge, alizadeh2021electroosmotic}. Moreover, nanoconfinement of the ionic solution is highly associated with the design and engineering of water treatment and transport systems\cite{qian2020nanoconfinement, cao2019water, wang2021efficient}. By forcing the ionic solution through nanoconfined space, ions can be filtered out when fresh water is produced\cite{mei2022simultaneous, cao2020single}. Ion concentration distribution in nanoconfined space and EDL is related to the performance and efficiency of the water treatment device or plants\cite{qian2020nanoconfinement, cao2021ozark, zhu2019structure, kong2017temperature}. Given the connection between nanoconfined ionic solution and various aforementioned applications, understanding the ion concentration distribution in nanospace is essential for the future development of energy storage, nanofluidic device, and water treatment technologies. 

Classical continuum models such as Poisson-Boltzmann equation have underwhelming accuracy in predicting ion concentration distribution under nanoconfinement\cite{freund2002electro, kekicheff1993charge} because nanoscale physical properties such as the effect of discrete solvent molecules and surface-ion interactions are not incorporated in those models\cite{qiao2004charge}. In the past two decades, molecular dynamics (MD) simulation has been the major tool for studying the ionic solutions in nanoconfinement as it models the physical interactions at molecular scale. MD simulation has significantly improved prediction accuracy on many properties of nanoconfined ionic solution such as ion and solvent concentration profile\cite{sendner2009interfacial, freund2002electro, qiao2003atypical, qiu2016ionic, robin2023ion, meidani2021titanium}, ion transport and velocity\cite{qiao2003ion, zhou2020field}, and charge inversion\cite{qiao2004charge, rojano2022effect} over continuum models. Despite its high accuracy, MD simulation has the disadvantage of being computationally expensive and resource-demanding. Additionally, to study the ionic distribution in nanochannel, a unique molecular system needs to be created and simulated for each combination of channel width and ion concentration. The trajectories from MD simulation have to be stored (occupying disk storage) and post-processed for the calculation of the desired properties. Therefore, building a surrogate model that makes instantaneous inference on specific physical properties of the ionic solution under different nanoconfinement configurations is of great value.

Machine learning and deep learning methods have seen astonishing development in the past decade due to the exponential growth in computational resources and data repositories. They are proven competent in modeling physical phenomena. For example, machine learning is used in identifying and modeling fluid mechanics\cite{brunton2020machine} and dynamical systems\cite{brunton2016discovering, cranmer2020discovering, meidani2023identification}, and solving mathematical problems\cite{davies2021advancing}. In the field of nanofluidics, graph neural network has been used to accelerate MD simulation\cite{li2022graph} and identify the phase of water\cite{moradzadeh2023top}, while convolutional neural network is used for predicting nanoconfined water density profile\cite{wu2022deep, santos2020modeling, lubbers2020modeling} and self-diffusion\cite{leverant2021machine} of Lennard-Jones fluids. 

In this work, we propose to use neural network (NN) as a surrogate model for the fast prediction of ion concentration profiles in nanochannels. There are several challenges in developing a surrogate NN model for this problem. Firstly, the ion concentration profile has a high-dimensional correlation with many properties of the nanochannel including the channel width, ion molarity, and ion type. Also, the shape of the ion concentration profile varies based on the bin size during data sampling. With those challenges, training an NN for end-to-end prediction of the entire concentration profile can be difficult and inflexible. To solve these issues, we treat the ion concentration profile in nanochannel as a probability distribution with respect to ion coordinates (details in the Method section) conditioned on nanochannel properties. We then train the NN as a parameterized approximator of the conditional cumulative density distribution of the ions in the nanochannel. Such a method is inspired by the fact that NNs are universal function approximators\cite{hanin2019universal, scarselli1998universal}, and it has achieved success in learning probability and cumulative distributions\cite{lu2020universal, magdon1998neural, liu2021density, rothfuss2019conditional}. NN trained in this manner learns a continuous distribution of ions in the nanochannel, which enables it to predict ion concentration profiles with arbitrary bin sizes. Results show that the NN can serve as a much faster surrogate model of MD simulation to accurately predict ion concentration profiles in nanochannels.

\section{Method}
\begin{figure*}[htb!]
\includegraphics[width=\linewidth]{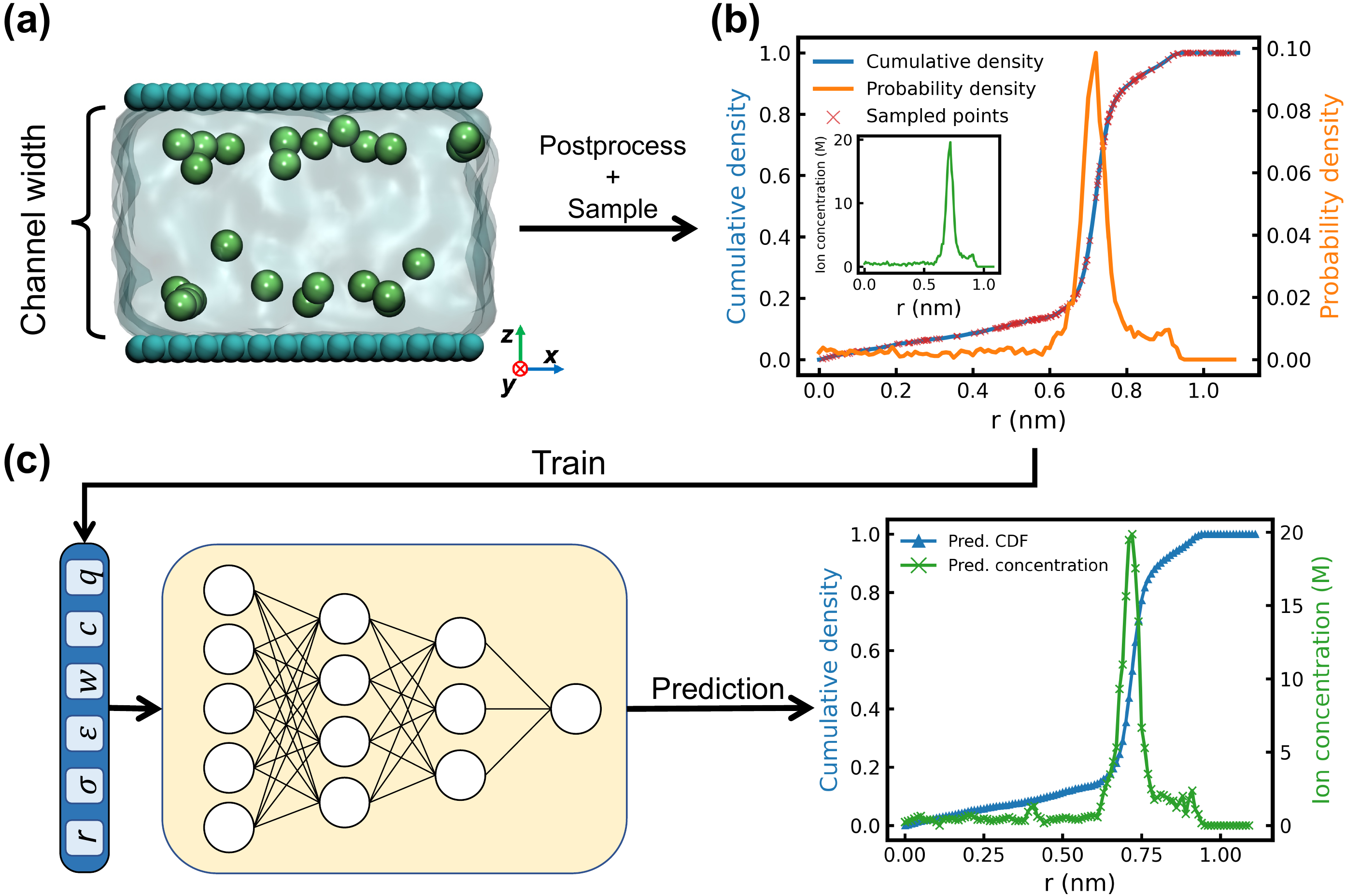}
\caption{Framework of predicting ion profiles in nanochannels via NN. \textbf{(a)} Data is generated by running MD simulation of nanoconfined ionic solution. Channel width, $w$, is the surface-to-surface distance between the two charged graphene walls (colored teal). In the shown system, the ions are \ce{Na^+} (colored green) with molarity of 2 M and $w$ is 2 nm. \textbf{(b)} shows the probability and cumulative density of ion distribution in the nanochannel. $r$ is the distance from the $z$-direction center of the nanochannel. It is noticeable that the ion concentration distribution shown by the inset is just a linear transformation of the probability distribution. Data points are sampled from the cumulative density and used for the training of NN. \textbf{(c)} The NN model takes 6 inputs and predicts the cumulative density of ion within $r$ nm of the channel center. The predicted cumulative density is then translated to ion concentration profile.}
\label{fig:pipeline}
\end{figure*}

The framework of this work (Fig.~\ref{fig:pipeline}) consists of three sections. Firstly, we perform molecular dynamics (MD) simulations of different ion-nanochannel systems. Secondly, data points are sampled based on the cumulative density function of ions in nanochannel to form a training dataset and an interpolation test dataset. Lastly, different machine learning models including NN are trained and further evaluated for their accuracy in predicting the ion concentration profile. 

\subsection{Molecular dynamics simulation}
The initial configurations of the ion-nanochannel systems are created using VMD\cite{humphrey1996vmd} and GROMACS\cite{bekker1993gromacs}. Each system (Fig.~\ref{fig:pipeline}a) contains a water box with ions sandwiched by two layers of graphene, which serve as the nanochannel walls. All systems have dimensions of 3.19 and 3.40 nm in the $x$ and $y$ direction, respectively. There are 832 carbon atoms in the simulation system. The surface-to-surface distance between the two graphene layers is the channel width denoted as $w$, ranging from 0.8 to 3.0 nm with the spacing of 0.1 nm ($w \in \{0.8, 0.9, \cdots, 3.0\}$, 23 variations). The number of water molecules in the nanochannel size of $3.19\times3.40\times w$ is controlled such that the water density is at 1 g/cm$^{-3}$. Ions of different molarity, $c$, are added to the system, where $c$ ranges from 0.8 to 3.6 M changing by 0.2 M ($c \in \{0.8, 1.0, \cdots, 3.6\}$, 15 variations). Five types of ions are simulated including \ce{Na^+}, \ce{Cl^-}, \ce{K^+}, \ce{Li^+}, and \ce{Mg^2+}. Each simulation system contains only one type of ion. To maintain charge neutrality, the same amount of opposite charges are evenly distributed to each carbon atom in the system, balancing the charge carried by the ions. In total, 1,725 simulations are run.

\begin{table}[]
\caption{The Lennard-Jones potentials and the charge of ions simulated in this work.}
\begin{ruledtabular}
\begin{tabular}{lllllll}
 & Na\cite{joung2008determination} & Cl\cite{joung2008determination} & Mg\cite{callahan2010solvation} & Li\cite{joung2008determination} & K\cite{joung2008determination} & C\cite{barati2011spatial}\\
\hline
$\sigma$ (\si{\angstrom}) & 2.1600 & 4.8305&2.1200 & 1.4094 & 2.8384 & 3.3900 \\
$\epsilon$ (kcal/mol) & 0.3526 & 0.0128 & 0.8750 & 0.3367 & 0.4297 & 0.0692\\
$q$ (e) & +1 & -1  & +2 & +1 & +1& - \\
\end{tabular}
\end{ruledtabular}
\label{tb:LJ} 
\end{table}

MD simulations are carried out using the LAMMPS\cite{plimpton1995fast} software. SPC/E\cite{mark2001structure} water model is used to simulate water molecules. The interatomic potentials consist of the Lennard-Jones (LJ) potentials and long-range Coulombic interaction, both with a cutoff distance of 10 $\si{\angstrom}$. The LJ potentials (Table~\ref{tb:LJ}) of ions 
 and graphene are adopted from the ref \cite{joung2008determination, callahan2010solvation, chan2010modelling} and ref\cite{barati2011spatial}, respectively. Other interatomic potentials are calculated by the arithmetic rule. Nos\'{e}--Hoover thermostat\cite{nose1984unified, hoover1985canonical} is used to maintain the temperature of the simulation system at 300 K. In each simulation, the system is firstly run under canonical (NVT) ensemble for 1 ns for the ions to reach equilibration positions. With the ensemble remains unchanged, the simulation is then run for another 1 ns while the trajectories are recorded every 5 ps. Periodic boundary condition is applied to all three dimensions. Post-processing of MD trajectories is conducted using the MDTraj\cite{McGibbon2015MDTraj} package.

\subsection{Ion concentration profile as probability distribution}
In this work, the ion concentration profile is regarded as the linear transformation of a conditional probability distribution (Fig.~\ref{fig:pipeline}b). Namely, the ion concentration is the conditional probability multiplied with molar concentration. When the material of nanochannel wall and the solvent molecule remains unchanged (graphene and water in our case), the probability of the $z$-coordinate of a nanoconfined ion is at $r$ nm away from the channel center follows the conditional probability density function (PDF):
\begin{equation}
    P(|z-o|= r \mid \sigma, \epsilon, w, c, q),
\label{eq:pdf}
\end{equation}
where $o$ is the $z$-coordinate of nanochannel center, $\sigma$ and $\epsilon$ are the LJ potentials of the ion, $w$ is the nanochannel width, $c$ is the molarity of ion in channel, and $q$ is the charge of the ion. If we denote the set including all conditions as $W = \{\sigma, \epsilon, w, c, q\}$, we get the conditional cumulative density function (CDF) as:
\begin{equation}
    F(|z-o| = r \mid W) = P(|z-o| \leq r \mid W) = \frac{N_{|z-o| \leq r}}{N_{\mathrm{total}}},
\label{eq:cdf}
\end{equation}
where $N_{|z-o| \leq r}$ is the number of ions having $z$-coordinate within $r$ nm from the nanochannel center, and $N_{\mathrm{total}}$ is the total number of ions in the system. If the conditional CDF is known, then the ion concentration in any interval $r_1 < |z-o| \leq r_2$ can be obtained simply by first calculating the conditional PDF:
\begin{multline}
    P(r_1 < |z-o| \leq r_2 \mid W) =\\
    F(|z-o| = r_2 \mid W) - F(|z-o| = r_1 \mid W)
\label{eq:interval}
\end{multline}
and then multiplying it by the molar concentration.

Since acquiring true conditional CDF shown by Eq.\ref{eq:cdf} is challenging, we propose to use a neural network parameterized by $\theta$, denoted as $F_{\theta}(\cdot)$, to approximate it:
\begin{equation}
    F_{\theta}(|z-o| = r \mid W) \approx F(|z-o| = r \mid W).
\end{equation}
Enough training data points should be sampled based on the true conditional CDF to ensure the expressivity of the neural network\cite{guhring2020expressivity, poole2016exponential}. In practice, we sample 2,000 data points using Eq.\ref{eq:cdf} from each simulation (Fig.~\ref{fig:pipeline}b). Among the 2,000 data points, half of them is sampled with uniformly chosen $r\in \{0, \frac{w}{2} + 0.1\}$ nm, which ensures the model learns that no ion should be present outside of the nanochannel. The other half is sampled with uniformly chosen $r\in \{\mathrm{max}(0, \frac{w}{2}-0.5), \frac{w}{2}\}$ nm. Such a choice of interval includes more data points within 5 $\si{\angstrom}$ of the nanochannel wall, thus helping the model to better learn the interfacial ion distribution. 3,450,000 data points are sampled in total from the 1,725 simulations. Each data point has features as $\{r, \sigma, \epsilon, w, c, q\}$ and labels as the conditional cumulative density $F(r \mid \sigma, \epsilon, w, c, q)$. Data points sampled from simulations with channel width $w\in \{1.6,  2.4,  2.8\}$ nm or with molarity $c \in \{1.4,  2.2,  3.0\}$ M are deposited into the test set, while the other ones into the training set. Overall, the training set contains 2,400,000 data points and the test set contains 1,050,000 data points. 

\subsection{Machine learning models and training}

In our work, we use NN to model the conditional CDF, $F(|z-o| = r \mid W)$. Each sample $i$ in $N$ training data is denoted as $\{r_i, W_i, F_i\}$ representing the descriptor $W_i$ and its corresponding density function value $F_i$ at position $r_i$. The input to the machine learning models is thus a six-dimensional list $\{r_i, W_i\} = \{r_i, \sigma_i, \epsilon_i, w_i, c_i, q_i \}$, which describes the conditions and position to predict (Fig.~\ref{fig:pipeline}). The output is $\tilde{F}_i$ which approximates the ground truth CDF $F_i$. An NN is developed as a surrogate model for ion concentration profile prediction. NN is based on fully-connected layers \cite{rumelhart1986learning, hinton2006reducing} and each layer is given:
\begin{equation}
    y=a(W^{(l)}x),
\end{equation}
where $x$ and $y$ are the input and output for each layer respectively, $W^{(l)}$ is the weight matrix at $l$-th layer, and $a$ is the element-wise nonlinear activation function. In our case, we used  ReLU\cite{maas2013rectifier} function for $a$ in all hidden layers and apply sigmoid in the output layer to guarantee the predicted cumulative density value is between 0 and 1. By stacking multiple fully-connected layers, NN can work as a universal function approximator \cite{lecun2015deep} which is expected to accurately predict the conditional ion profiles. The objective of the NN is to minimize the MSE between the predictions $\{\tilde{F}_i\}_{i \in N}$ and the ground truth $\{{F}_i\}_{i \in N}$. The NN is implemented with five hidden layers and the number of units in each hidden layer is $\{1024, 512, 256, 128, 32\}$. We train the NN using Adam optimizer \cite{kingma2014adam} with a learning rate of 0.005 for 100 epochs.

To demonstrate the superior performance of NN, we benchmarked the prediction of ion profiles using another machine learning model, extreme gradient boosting (XGBoost). XGBoost is a gradient-boosted decision tree (GBDT) machine learning method and has demonstrated superior and robust performance in many applications \cite{chen2016xgboost}. Unlike the NN, XGBoost is a non-parametric machine learning model without strong assumptions about the form of the mapping function from input to output. The model is based on decision trees which predict the labels via a series of if-then-else questions (levels) to the input features. An optimal decision tree estimates the minimum number of levels required to make a prediction during training. GBDT is an ensemble method for decision trees that combines multiple models to boost performance \cite{mason1999boosting, ke2017lightgbm}. Specifically, GDBT leverages the error residual of the previous decision tree to train the next and ensembles all the decision trees by a weighted summation of the outputs of the models. For a GBDT $\tilde{F}_M$ with $M$ decision trees, the ensembled result is given:
\begin{equation}
    \tilde{F}_M(z, W) = \sum_m^M \gamma_m T_m(r, W),
\end{equation}
where $T_m$ denotes a single decision tree. And the weighted term $\gamma_m$ is calculated by:
\begin{equation}
    \gamma_m = \argmin_\gamma \sum_i^N \ell (F_{i}, \tilde{F}_{m-1}(r_i,W_i) + \gamma T_{m}(r_i, W_i) ),
\end{equation}
where $\ell$ is a mean square error (MSE) loss $\ell(y, \tilde{y}) = (y - \tilde{y})^2$ that measures the prediction accuracy. XGBoost is a scalable and distributed implementation of GDBT, which builds and trains decision trees in parallel. In our case, we set the level of decision trees as 15 and the $L_2$ regularization coefficient as 5 to mitigate overfitting. Other hyperparameters follow the default settings in the XGBoost package\cite{chen2016xgboost}. 

\section{Result and Discussion}
\subsection{Prediction of ion concentration profile}

\begin{figure}[htb!]
\includegraphics[width=\linewidth]{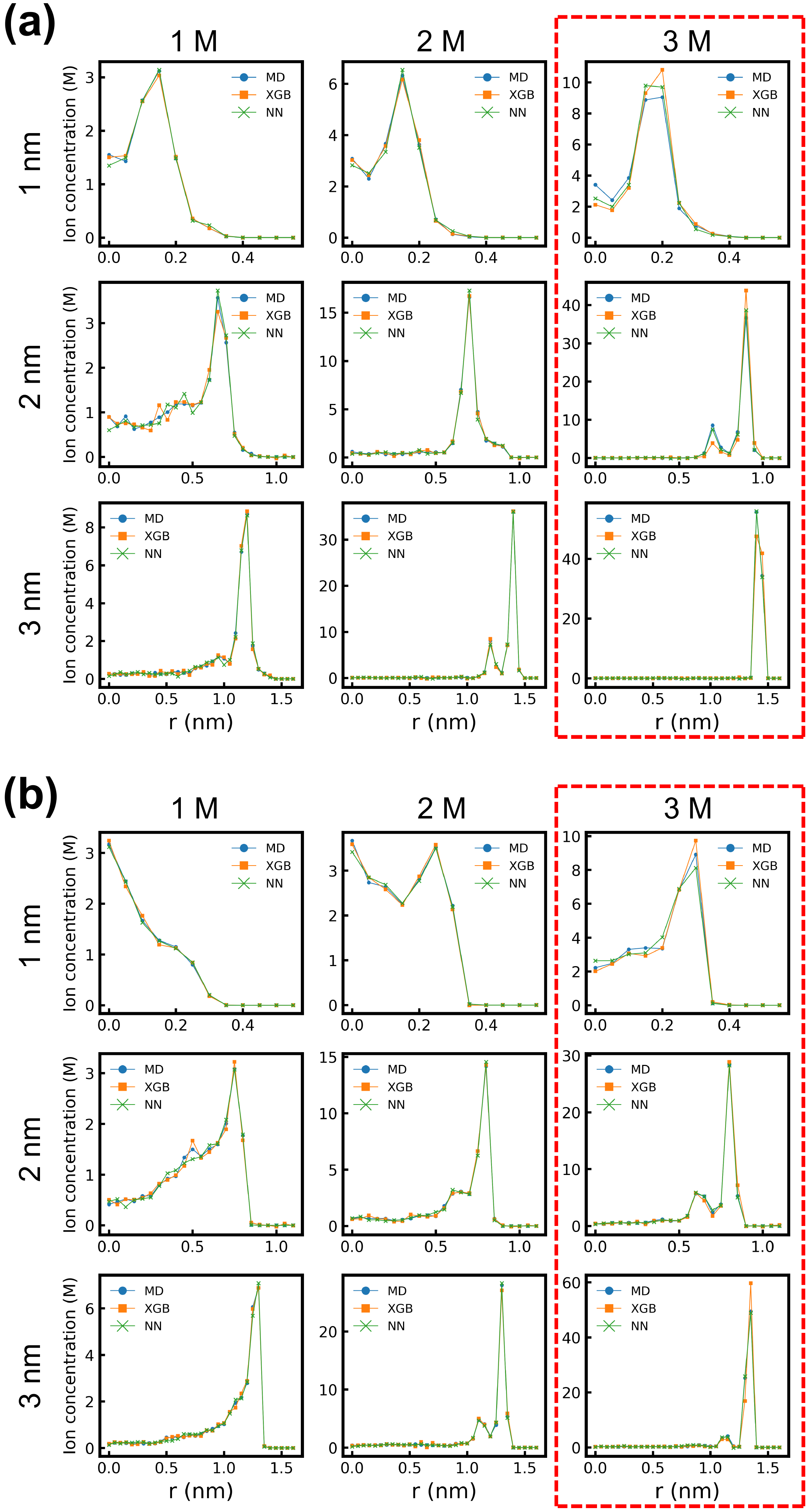}
\caption{The comparison between ion concentration profiles predicted by MD simulation, neural network, and XGBoost model when channel width ranges from 1 to 3 nm and molarity ranges from 1 to 3 M. \textbf{(a)} shows the results of \ce{Na^+} and \textbf{(b)} shows the results of \ce{Cl^-}. Results enclosed by the red dashed lines are included in the interpolation test set, while the others are included in the training set. $r$ represents the distance to the center of nanochannel.}
\label{fig:profile}
\end{figure}

\begin{figure*}[htb!]
\includegraphics[width=\linewidth]{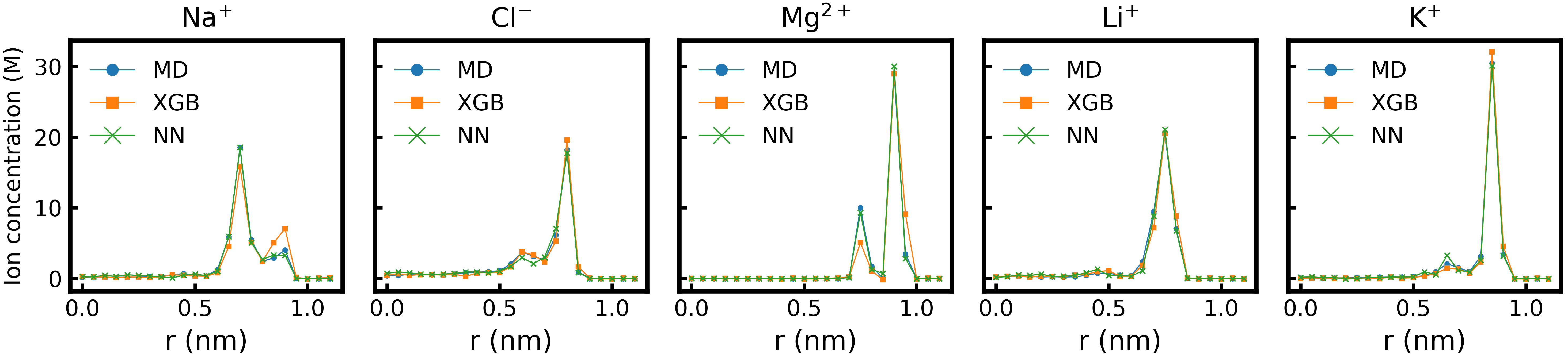}
\caption{Comparison between model predicted ion concentration profile and MD simulation results for five different ions. The channel width is 2 nm and molarity is 2.2 M, which is a configuration included in the interpolation test set. Bin size is set as 0.05 nm.}
\label{fig:ions}
\end{figure*}

To examine the performance of the NN and XGBoost models, we compare their predicted ion concentration profiles with the MD simulation results. Fig.~\ref{fig:profile} shows the ion concentration profile predicted using the three methods with different system configurations (combination of channel width and molarity), where the bin size is set as 0.05 nm. Plots enclosed by the red dashed lines have either channel width or molarity in the interpolation test set, while the other plots are results from the training set. For \ce{Na^+} (Fig.~\ref{fig:profile}a), both models achieve close to perfect prediction for configurations in the training set. The location of the model-predicted concentration peaks fits perfectly with the MD simulation results. NN demonstrates more accurate interpolation capability than XGBoost as XGBoost has higher errors in predicting the peak concentration when molarity is 3 M. For \ce{Cl^-} (Fig.~\ref{fig:profile}b), XGBoost shows a tendency of overpredicting the peak concentration when the molarity is at 3 M. A single layer of \ce{Cl^-} occurs in the configuration [1 nm, 1 M], which is different from the other configurations where ion concentration peak occurs closer to the nanoconfinement wall. Both models accurately capture such a physical phenomenon. One small error made by NN is that it fails to predict the secondary concentration peak located at 0.5 nm from the channel center in the configuration [2 nm, 1 M]. In general, Fig.~\ref{fig:profile} demonstrates that NN is capable of accurately predicting the ion concentration profile by learning the CDF of ion in nanochannel, and NN interpolates better on unseen configurations compared with XGBoost. 

Another variable besides channel width and molarity is the ion type. Fig.~\ref{fig:ions} shows the comparison between model-predicted ion concentration profiles and MD simulation results for five different ions. The channel width is 2 nm, the molarity is 2.2 M, and the bin size is 0.05 nm for all subplots in Fig.~\ref{fig:ions}. It can be observed that the location of peaks predicted by both NN and XGBoost is correct compared with MD results. For \ce{Na^+}, XGBoost underpredicts the concentration of the primary peak while largely overpredicting the secondary peak. XGBoost also significantly underpredicts the concentration of the secondary peak of \ce{Mg^{2+}}. Despite minor prediction error found on the secondary peaks of \ce{Cl^-} and \ce{K^+} profiles, NN is shown to be a more accurate model than XGBoost in this task. Ion concentration profiles of all configurations are available on the \href{https://github.com/zcao0420/IonNet/tree/main/profile_plots}{GitHub page} of this work. Fig.~\ref{fig:ions} shows that NN bears the potential as a generalizable predictor of ion concentration profile regardless of the type and charge of the ions.

\subsection{Error analysis and inference time}

\begin{figure}[htb!]
\includegraphics[width=\linewidth]{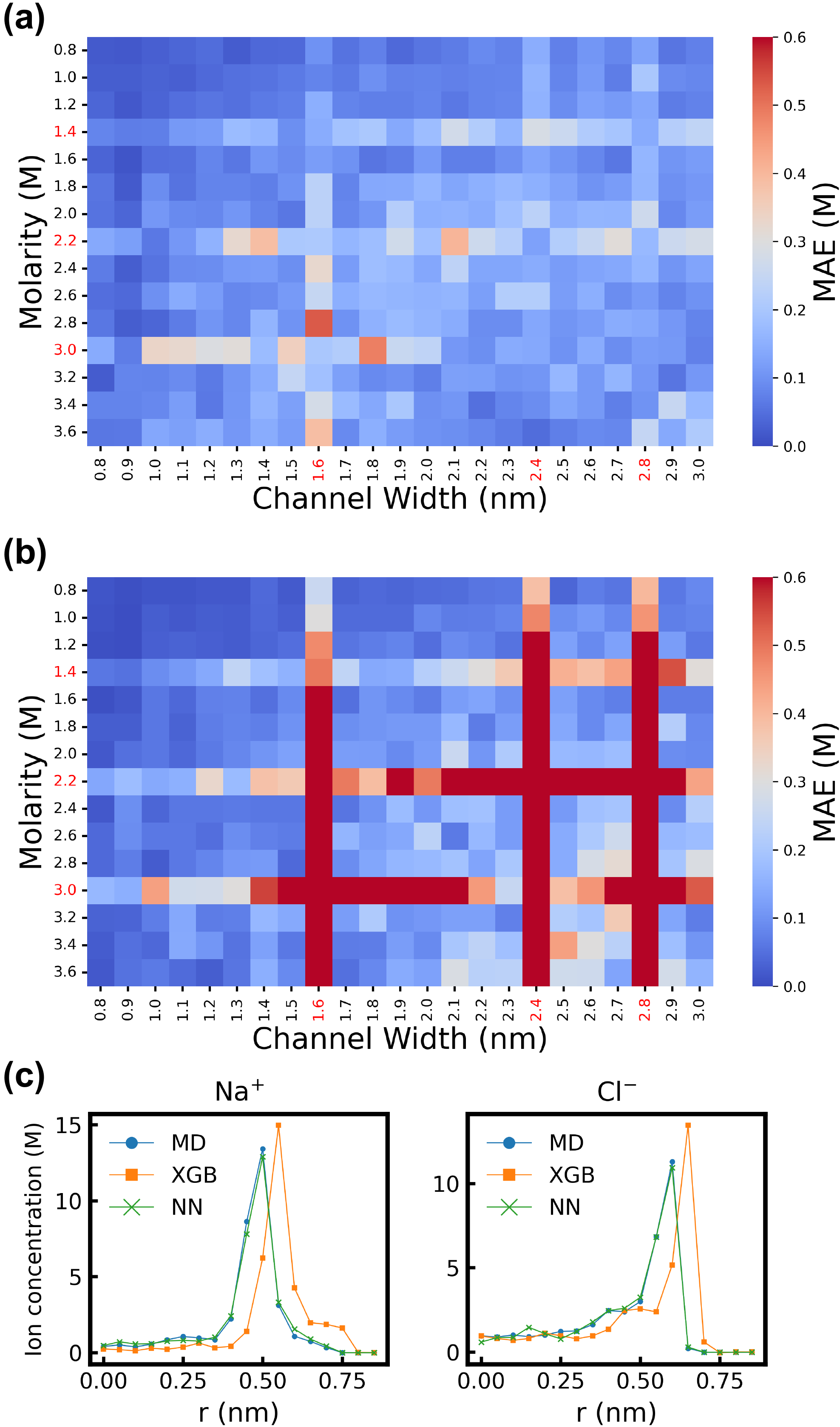}
\caption{Mean absolute error of \textbf{(a)} NN and \textbf{(b)} XGBoost predicted \ce{Na^+} concentration profile in all system configurations. MAE is calculated based on the profiles with bin size of 0.05 nm. The highest MAE of XGBoost in the dark red blocks is 4.92 M. The maximum value of the color bar is truncated such that the MAE of NN can be distinguished. The tick labels of channel width and molarity in the test set are colored red. \textbf{(c)} shows examples of failed interpolation of XGBoost for the configuration [1.6 nm, 2.2 M] due to overfitting.}
\label{fig:error}
\end{figure}

To thoroughly evaluate NN and XGBoost as a surrogate model to MD simulation, we benchmarked both models on their prediction accuracy and inference time. The mean absolute error (MAE) of \ce{Na^+} concentration profile prediction is the first metric of accuracy. Fig.~\ref{fig:error}a and \ref{fig:error}b visualize the MAE heatmap of NN and XGBoost, respectively, for all simulated configurations. The MAE is calculated based on concentration profiles with bin size of 0.05 nm. The tick labels of the channel width and molarity included in the interpolation test set are colored red. Despite a slight increase of MAE when interpolating, NN achieves high prediction accuracy for all configurations. The highest MAE of NN is 0.53 M in the configuration [1.6 nm, 2.8 M]. On the other hand, the prediction accuracy of XGBoost suffers from overfitting. When interpolating \ce{Na^+} concentration profile for 1.6, 2.4, 3.0 nm channel width and 1.4, 2.2, 3.0 M molarity, MAE of XGBoost drastically increases. The highest MAE of XGBoost is 4.92 M, which occurs in the configuration [1.6 nm, 3.6 M]. The maximum value of the color bar of the heatmap is truncated at 0.6 M so that low MAE is more visible. Examples of XGBoost overfitting \ce{Na^+} and \ce{Cl^-} concentration profiles in configuration [1.6 nm, 2 M] are shown in Fig.~\ref{fig:error}c. For both \ce{Na^+} and \ce{Cl^-}, the concentration peaks are mispredicted by XGBoost to be 0.05 nm away from the MD simulation results, leading to exceptionally high MAE. On the other hand, overfitting has a trivial effect on the prediction accuracy of NN. When channel width is 2.4 nm, no obvious MAE increase is shown in Fig.~\ref{fig:error}a. 

\begin{table}[b!]
\caption{Evaluation of the interpolation error and inference time of NN and XGBoost. All metrics are calculated given the bin size of 0.05 nm. Both MAE and peak deviation are averaged over all configurations in the interpolation set. Inference time measures the total time to generate ion concentration profiles for all 1,725 configurations (5 ions, 23 channel width, 15 molarity). The inference time of MD does not include the time used to load molecular trajectories. The inference time reported is averaged over 6 runs, with the standard deviation in parentheses. The lower value of each metric is the better model.}
\begin{ruledtabular}
\begin{tabular}{llll}
 & NN & XGBoost & MD\\
\hline
Interp. MAE (M) & 0.1899 & 1.2764&-\\
Interp. peak deviation (nm) & 0.0019 & 0.0314 & -\\
Inference time (s) & 0.80(0.10) & 2.56(0.11)  & 11.72(0.42) \\
\end{tabular}
\end{ruledtabular}
\label{tb:error} 
\end{table}

Two metrics, MAE and peak deviation are calculated to quantitatively evaluate the advantage of NN in interpolation (Table~\ref{tb:error}). MAE measures the prediction error of the whole ion concentration profile. Peak deviation measures how well the shape of the predicted ion concentration profile is aligned with the MD simulation result. The lower the value for each metric, the better the model is expected in interpolation. The interpolation MAE of NN is only 14.9\% of that of XGBoost. Moreover, the interpolation peak deviation of NN is 0.0019 nm, which is only 6.1\% of that of XGBoost. Both metrics prove that NN does not suffer from overfitting as XGBoost does, and is capable of accurately interpolating ion concentration profiles under different nanoconfinement conditions. 

Using machine learning as a fast surrogate model for MD simulation is the major motivation for this work. To evaluate the improvement in computation speed, we benchmark the inference time of NN, XGBoost, and MD simulation. Here, we define the inference time to be the total time taken to predict the ion concentration profiles of all 1,725 system configurations (5 ions, 23 channel widths, 15 molarities) with bin size as 0.05 nm. For MD simulation, we only consider the postprocessing time of the molecular trajectories using MDTraj\cite{McGibbon2015MDTraj} and NumPy\cite{harris2020array, van2011numpy}. The time of loading molecular trajectories is not included for a fair comparison. The benchmark is conducted with an Intel Core i7-8086K CPU for all models (Table~\ref{tb:error}). Averaged over 6 runs, NN can predict all 1,725 ion concentration profiles in 0.8 s, whereas XGBoost takes 2.56 s and processing MD trajectories takes 11.72 s. NN is approximately 3.2$\times$ faster than XGBoost and 14.7$\times$ faster than processing MD simulation trajectories. The inference time benchmark demonstrates that NN is a fast surrogate model for MD simulation. It is noteworthy that running a single MD simulation takes about 8 hours on average using 6 CPU cores, indicating that running all 1,725 simulations takes approximately 82,800 core-hours. If we compare the inference time of NN to the time required to run MD simulation, NN is orders of magnitude faster. Considering the size of NN model is 2.8 MB, which is much smaller than XGBoost model (71.1 MB) and MD simulation trajectories (32.1 GB), NN is also more storage-efficient than the other two.

\subsection{Sampling with different bin sizes}

\begin{figure}[htb!]
\includegraphics[width=\linewidth]{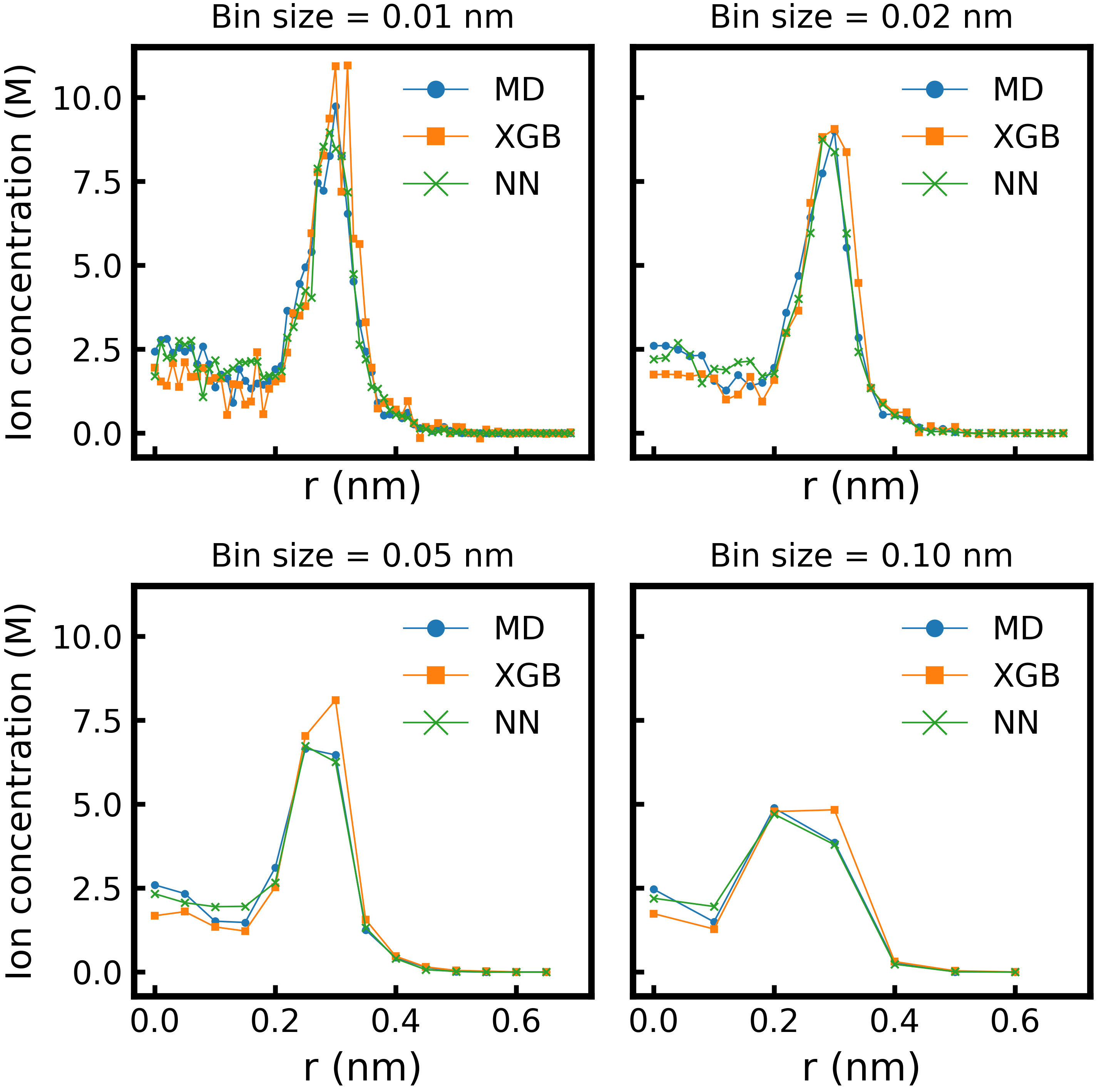}
\caption{Predicted \ce{Na^+} concentration profile with bin size ranging from 0.01 to 0.1 nm. The shown configuration is [1.2 nm, 2.2 M], which is included in the interpolation set.}
\label{fig:binsize}
\end{figure}

One advantage of training a NN to learn the CDF instead of the whole ion concentration profile is the flexibility of sampling. By sampling on the CDF and converting probability to ion concentration, the NN can predict ion concentration profiles with arbitrary bin sizes. Fig.~\ref{fig:binsize} shows the prediction of \ce{Na^+} concentration profile in the configuration [1.2 nm, 2.2 M] with bin size ranging from 0.01 to 0.1 nm. Both NN and XGBoost successfully model the shape of the profile regardless of the bin size: a primary peak near the confinement wall and a secondary peak at the channel center. When bin size is as fine as 0.01 nm, the concentration profile near the channel center is noisy because of less ion presence. Given the stochastic nature of MD simulation, which is used to generate the training data, the minor prediction error near the channel center with such a fine bin size is expected. The effect of stochasticity reduces as the bin size increases, resulting in more accurate predictions by NN and XGBoost. 

\section{Conclusion}
In this work, we show the viability of using a neural network (NN) as a surrogate model for MD simulation in predicting the ion concentration profile in the nanochannel. We first model the ion concentration profile in nanochannel as a probability distribution conditioned on the configuration of nanochannel systems, such as LJ potentials of the ion, channel width, and molarity of ion. A NN is then trained to predict the conditional cumulative density probability (CDF) distribution of ions given the system configuration. The predicted CDF can be linearly transformed back to the concentration profile as the final output of the NN model. 1,725 MD simulations (5 types of ions, 23 variations of channel width, 15 variations of molarity) are run to generate training and test data for the NN model.

We benchmark NN against another machine learning model, XGBoost, to compare their interpolation ability. The two metrics we used are the mean absolute error (MAE) of the whole interpolated ion concentration profile, and the peak deviation from the MD simulation concentration profile. On average, NN achieves 85.1\% lower mean absolute error and 93.9\% lower peak deviation when interpolating \ce{Na^+} concentration profiles compared with the XGBoost model. Such a benchmark demonstrates the superior ability of NN to interpolate ion concentration profiles given unseen system configurations. Further, we record the inference time taken by the NN and XGBoost models and compared that to the time taken to post-process MD simulation trajectories. To predict 1,725 ion concentration profiles, NN takes an inference time of 0.80 s, which is approximately 3.2$\times$ and 14.7$\times$ faster than the XGBoost and processing MD trajectories, respectively. At last, we show that NN can flexibly predict ion concentration profile with any bin size. The accuracy of NN prediction increases with larger bin sizes because of the reduced effect of stochasticity. In conclusion, NN is a fast, flexible, and most importantly, accurate surrogate model to predict ion concentration profiles under nanoconfinement.

\begin{acknowledgments}
We wish to acknowledge the computational resources provided by the Pittsburgh Supercomputing Center (PSC). Z.C. acknowledges the funding from the Neil and Jo Bushnell Fellowship in Engineering.
\end{acknowledgments}

\section*{Data Availability Statement}

The training data used for machine learning model training are openly available on GitHub (\href{https://github.com/zcao0420/IonNet}{https://github.com/zcao0420/IonNet}).

\bibliography{main}

\end{document}